\documentclass[10pt,letterpaper,twocolumn]{article}
\usepackage[T1]{fontenc}
\usepackage{newtxtext}
\usepackage{helvet}
\usepackage{courier}
\usepackage[margin=0.75in,columnsep=0.25in]{geometry}
\usepackage[hyphens]{url}
\usepackage{graphicx}
\usepackage{float}
\usepackage{natbib}
\usepackage{caption}
\usepackage[hidelinks]{hyperref}
\usepackage{microtype}
\usepackage{placeins}
\usepackage{amsmath}
\usepackage{amssymb}
\usepackage{booktabs}
\usepackage{multirow}
\usepackage{makecell}
\usepackage{threeparttable}

\title{\textbf{TrainSDC: Characterizing and Mitigating Silent Data Corruption in Large Language Model Training}}
\author{
    Zhipeng Xia$^{1}$,
    Haotian Xu$^{1}$,
    Siyu Yun$^{1}$,
    Liqi Lin$^{1}$,\\
    Hu Liu$^{2}$,
    Yu Li$^{1,*}$,
    Cheng Zhuo$^{1}$\\[0.5em]
    \small $^{1}$Zhejiang University \quad $^{2}$Huawei\\
    \small $^{*}$Corresponding author: yu.li.sallylee@gmail.com
}
\date{}

\begin{document}

\maketitle

\begin{abstract}

LLM training is increasingly vulnerable to silent data corruption (SDC), yet existing protection methods largely treat Transformer computations uniformly because their vulnerability remains poorly understood. We present the first systematic characterization of SDC vulnerability across major computation interfaces in both the forward and backward passes of Transformer training. Our analysis reveals two distinct error propagation mechanisms: forward-pass vulnerability is highly location dependent, with faults on the Q/K path producing persistent training deviations, whereas backward-pass vulnerability is largely governed by gradient exponent distributions rather than computation locations. Motivated by these observations, we propose TrainSDC, a characterization-guided protection framework consisting of Q/K-path recomputation, residual-gain monitoring, and exponent-aware gradient scaling. Experiments on Llama 3.2-1B and Qwen3-0.6B show that TrainSDC maintains training behavior close to fault-free execution under both sparse and dense fault injection while introducing only 1.65\%–6.76\% runtime overhead.

\end{abstract}

\section{Introduction}
In LLM training, silent data corruption (SDC) occurs when a hardware fault
produces an incorrect result without triggering an exception or error signal.
Because the corrupted value is consumed as valid, it can flow through
activations and gradients into parameter updates even when the loss and other
global signals show no obvious anomaly.
The damage may become visible only later as a loss spike or failed convergence,
or remain hidden while degrading final model quality
\citep{dixit2021silent,ma2025understanding}. Training larger models typically
requires more accelerators and longer runs, increasing the chance that SDC
affects training. At the scale of Gemini, the team estimated that SDC events
could affect training every one to two weeks
\citep{team2023gemini}.

Without a direct alarm, assessing SDC risk requires understanding how a
corrupted value propagates and whether its effect disappears or persists in
later updates. A large immediate deviation is not necessarily the most harmful:
some errors fade as training proceeds, whereas others bias later updates and
leave persistent damage. A useful characterization must therefore consider both
where a fault enters the training computation and how its effect evolves over
time. Studies using faulty production hardware analyze SDC at the levels of
submodule computation, a single optimizer step, and an extended training period
\citep{ma2025understanding}. Large injection campaigns characterize the
outcomes of transient faults in LLM training \citep{yu2025exploring}.
Instruction- and RTL-level studies vary injection sites, fault patterns, and
numerical formats
\citep{altenbernd2026exploring,tyagi2026llmprism}. However, these studies do not
jointly compare the major Transformer components in the forward and backward
passes or distinguish short-lived effects from damage that remains at the end
of training.

This incomplete understanding also limits protection design. Existing
defenses infer corruption from aggregate training signals
\citep{altenbernd2026exploring,yu2025exploring}, check selected computations
\citep{liang2025attnchecker}, identify faulty devices
\citep{lei2026safeguarding}, or duplicate execution
\citep{park2026sparetrain}. These methods either operate at coarse granularity,
cover only selected computations, or incur the cost of redundant execution.
Without a detailed characterization of error propagation, it remains unclear
where direct verification is needed and where lightweight monitoring is
sufficient.

We therefore conduct a systematic fault-injection study of pre-norm LLMs. We
inject faults at the outputs of the major components in a Transformer block
during both forward and backward passes. We measure the largest loss change and
the remaining change at the end of training to distinguish faults that fade
from those that cause lasting damage. Based on the different ways faults
propagate in the forward and backward passes, we design \textit{TrainSDC} with
a separate protection method for each pass.

Our contributions are threefold:
\begin{itemize}
  \item To our knowledge, we present the first systematic characterization of
  SDC vulnerability across all major modules in a Transformer block
  during both forward and backward passes. 
  \item We find that faults in Q/K path outputs cause more persistent
  forward-pass damage, whereas faults in value, attention-output, and MLP
  computations cause larger but shorter-lived loss changes. Backward-pass
  vulnerability varies less across components, and fault amplification depends
  on the exponent distribution of gradients.
  \item We develop \textit{TrainSDC}, which checks Q/K operations through
  recomputation, monitors residual connections, and re-executes flagged steps
  in the forward pass. For the backward pass, it selects a power-of-two loss
  scale from clean gradient statistics.
\end{itemize}

On Llama~3.2-1B and Qwen3-0.6B, TrainSDC mitigates both sparse and dense faults with limited overhead. It maintains results close to fault-free training under sparse faults and preserves convergence under dense faults causing unprotected training to diverge. Overall, TrainSDC restores training outcomes to a level close to fault-free training.

\section{Related Work}

\noindent\textbf{Characterizing SDC in LLM training.}
Prior work studies SDC through production-system measurements and controlled
fault injection. Fleet-scale measurements document the prevalence and diversity
of SDC in production hardware \citep{dixit2021silent}. Experiments on unhealthy
training nodes analyze real-world SDC at three scales: attention and FFN
outputs, a single optimizer step, and an extended training period
\citep{ma2025understanding}. These experiments preserve realistic hardware
behavior, but do not control fault location and timing or separate the
individual computations within attention and FFN. Controlled studies inject
faults at GPU matrix-multiply instructions
\citep{altenbernd2026exploring}, run large transient-fault campaigns
\citep{yu2025exploring}, or combine RTL-level simulation with training-level
injection to study permanent faults \citep{tyagi2026llmprism}. Our work differs
in granularity and comparison scope: under the same fault settings, we inject
faults separately at the outputs of individual Transformer computations and
into their corresponding backward gradients. This allows us to compare
vulnerability across locations and passes and distinguish short-lived from
lasting effects.

\noindent\textbf{Training-time SDC protection.}
Some methods use training-level statistics to detect faults. Altenbernd et al.
identify harmful updates from changes in the magnitude of AdamW parameter
updates and the global gradient norm \citep{altenbernd2026exploring}. LLMFT
instead uses heuristic training features as input to a learned fault detector
\citep{yu2025exploring}. These methods do not check individual Transformer
computations. ATTNChecker applies algorithm-based fault tolerance (ABFT) to
detect and correct extreme errors in attention
\citep{huang1984abft,liang2025attnchecker}. SpareTrain reduces the overhead of
complete dual modular redundancy by reusing activation checkpointing and idle
GPU time \citep{park2026sparetrain}. ATTNChecker protects attention, whereas
SpareTrain duplicates the full computation.
Other methods focus on distributed training. PAFT handles gradient-aggregation
errors \citep{tangidentifying}; AEGIS detects SDC online and identifies faulty GPUs
\citep{lei2026safeguarding}; and ByteRobust diagnoses failures and accelerates
recovery \citep{wan2025robust}. These methods address gradient aggregation,
faulty devices, or job recovery rather than local Transformer computations.

In contrast, we compares SDC vulnerability across the major computations
in a Transformer block and separately analyzes the forward and backward passes.
TrainSDC uses this characterization to protect the two passes differently
without continuously duplicating the full training computation.

\section{Fault Characterization}
\label{sec:fault-characterization}

\begin{figure*}[t]
  \centering
  \includegraphics[width=0.93\textwidth]{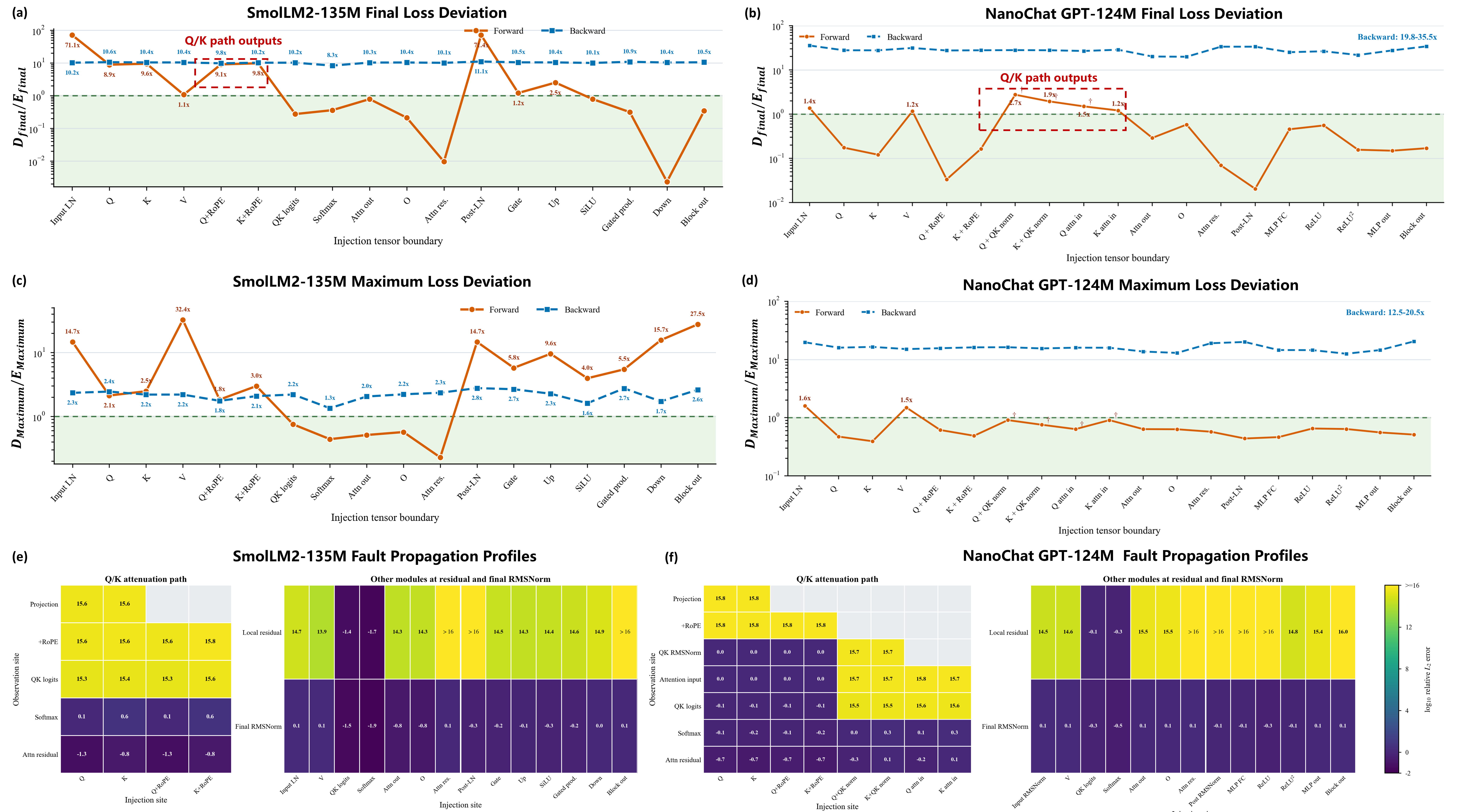}
  \caption{Characterization results for SmolLM2-135M and NanoChat-GPT-124M.
  Panels (a) and (b) report the normalized final loss deviation, panels (c)
  and (d) report the normalized maximum loss deviation, and panels (e) and
  (f) report fault propagation using relative $L_2$ error.
}
  \label{fig:2}
\end{figure*}

This section characterizes how transient computation errors affect large
language model training. We examine how their effects vary across module
outputs and between forward and backward stage, and trace whether the
errors are amplified, attenuated, or retained. These
results provide the basis for the protection design presented in the next
section.

\subsection{Fault Model and Metrics}
\label{sec:fault-model}

\paragraph{Fault type.}
Persistent failures in storage are generally easier to detect because storage
systems commonly use error-correcting codes and related integrity checks.
In contrast, transient errors in computation can produce an incorrect result without leaving a persistent state and are therefore harder to detect. A transient
error affects only the current tensor computation, and neither the corrupted
value nor its bit state is retained by the next execution. We focus on transient errors in arithmetic computation and do not model persistent storage failures.

\paragraph{Injection location.}
We inject errors into module activation outputs during the forward pass and their
corresponding output gradients during the backward pass. For an activation
$x$, the backward target is $g_x=\partial L/\partial x$. In each target module,
we randomly select $m$ elements and flip four randomly
selected bits in each element. All module outputs use the same random seeds,
and injection is performed in the original data type. Parameters,
optimizer states, and stored training data are not modified.

\paragraph{Loss-deviation metrics.}
The loss-deviation metrics separate the largest deviation during training from
the deviation at the final step. Let $t_f$ denote the first faulty step and $T$
the final step:

\begin{align}
D_{\mathrm{Maximum}} &\triangleq
  \max_{t_f\leq t\leq T}
  \left(L_{\mathrm{fault}}(t)-L_{\mathrm{clean}}(t)\right),\\
D_{\mathrm{final}} &\triangleq
  L_{\mathrm{fault}}(T)-L_{\mathrm{clean}}(T).
\end{align}

Repeated clean trajectories estimate nondeterministic training variation.
$E_{\mathrm{final}}$ is the largest pairwise final-loss difference among clean
runs, while $E_{\mathrm{Maximum}}$ is the largest pairwise loss difference at
any aligned step. Figures report
$D_{\mathrm{final}}/E_{\mathrm{final}}$ and
$D_{\mathrm{Maximum}}/E_{\mathrm{Maximum}}$. A value above one exceeds every
observed clean-to-clean difference under the corresponding metric.

\subsection{Experimental Design}
\label{sec:characterization-design}


\paragraph{Experiment 1: training vulnerability.}
Experiment 1 measures training vulnerability at forward
and backward stage separately. We select SmolLM2-135M and NanoChat-GPT-124M as the experimental models because
their modern Transformer architectures are representative of compact language
models. In the injection experiment, SmolLM2 targets
layer 20 and runs from steps 1222 to 2035, whereas NanoChat-GPT targets layer 5
and runs from steps 955 to 1907. Each run modifies rank 0, corrupts
$100{,}000$ elements at $1\%$ of training steps, and uses the shared random
selection described above. Repeated no-fault runs establish
$(E_{\mathrm{final}},E_{\mathrm{Maximum}})$ as
$(0.000456,0.009299)$ for SmolLM2 and
$(0.000389,0.003153)$ for NanoChat-GPT.

\paragraph{Experiment 2: propagation diagnostic.}
Experiment 2 traces the propagation of the injected fault through downstream
modules. The experiment uses the same target layers,
element selection, and bit selection as Experiment 1. Each run injects one
module output and records the resulting changes at downstream attention,
residual, multilayer-perceptron, normalization, and language-model output. For a monitored module output tensor $x$, the propagation metric is
\begin{equation}
  R_x =
  \frac{\lVert x_{\mathrm{fault}}-x_{\mathrm{clean}}\rVert_2}
       {\lVert x_{\mathrm{clean}}\rVert_2}.
\end{equation}
The numerator measures the difference between the faulty and clean tensors,
while the denominator normalizes this difference by the magnitude of the clean
tensor. Figures~\ref{fig:2}(e) and (f) report $\log_{10}R_x$, where a positive
value indicates that the difference exceeds the clean tensor magnitude. A larger value therefore indicates a more pronounced effect of the fault.


\subsection{Characterization Results}
\label{sec:characterization-observations}


\paragraph{Observation O1: Forward sensitivity is path-dependent, whereas backward sensitivity is broad.}
Across both models, the final loss is most sensitive to faults in the Q/K path outputs, namely the Q and K tensors consumed by the attention operator, together with the input and post-attention normalization outputs. This ranking is architecture-invariant: although NanoChat-GPT applies Q/K RMSNorm after the projections and thus absorbs projection-stage errors, the Q/K path outputs remain the most vulnerable interface in both architectures.

Figure~\ref{fig:2} further shows that peak and final loss deviations follow different rankings. Faults in the V, normalization, MLP, and block outputs produce the largest immediate loss perturbations, but these disturbances are gradually attenuated by normalization and subsequent clean updates. In contrast, faults in the Q/K path outputs produce only modest immediate loss perturbations while leading to the largest final loss deviations, indicating that they silently alter the optimization trajectory rather than causing conspicuous short-term failures.

Backward injection exhibits much weaker module differentiation. Most monitored gradients lead to similarly large training deviations, suggesting that backward vulnerability is not dominated by individual Transformer modules but instead requires protection with broad coverage.

\textbf{Design implication.}
Forward protection should distinguish computations according to their propagation characteristics. Q/K-path faults are difficult to detect from downstream signals despite their lasting impact, whereas backward protection should employ a unified mechanism rather than module-specific protection.


\paragraph{Observation O2: Q/K-path faults are weakly observable, whereas most other forward faults remain detectable after propagation.}

Figures~\ref{fig:2}(e) and (f) show that faults in the Q/K path outputs are strongly attenuated by the softmax operation and subsequent residual addition. Consequently, even substantial corruption often results in only modest downstream loss perturbations and can escape both loss-based and activation-magnitude monitoring, although the altered attention allocation continues to influence subsequent parameter updates.

In contrast, faults originating from the V path, normalization, attention output, and MLP remain visible as abnormal magnitude changes at the following residual writeback. These faults therefore preserve observable signatures after propagation, making them amenable to lightweight runtime monitoring.

\textbf{Design implication.}
Protection should directly verify computations whose faults become weakly observable after propagation, while computations that preserve detectable propagation signatures can instead be protected through lightweight monitoring.

\FloatBarrier
\section{Protection Design}

\begin{figure*}[t]
  \centering
  \includegraphics[width=0.93\textwidth]{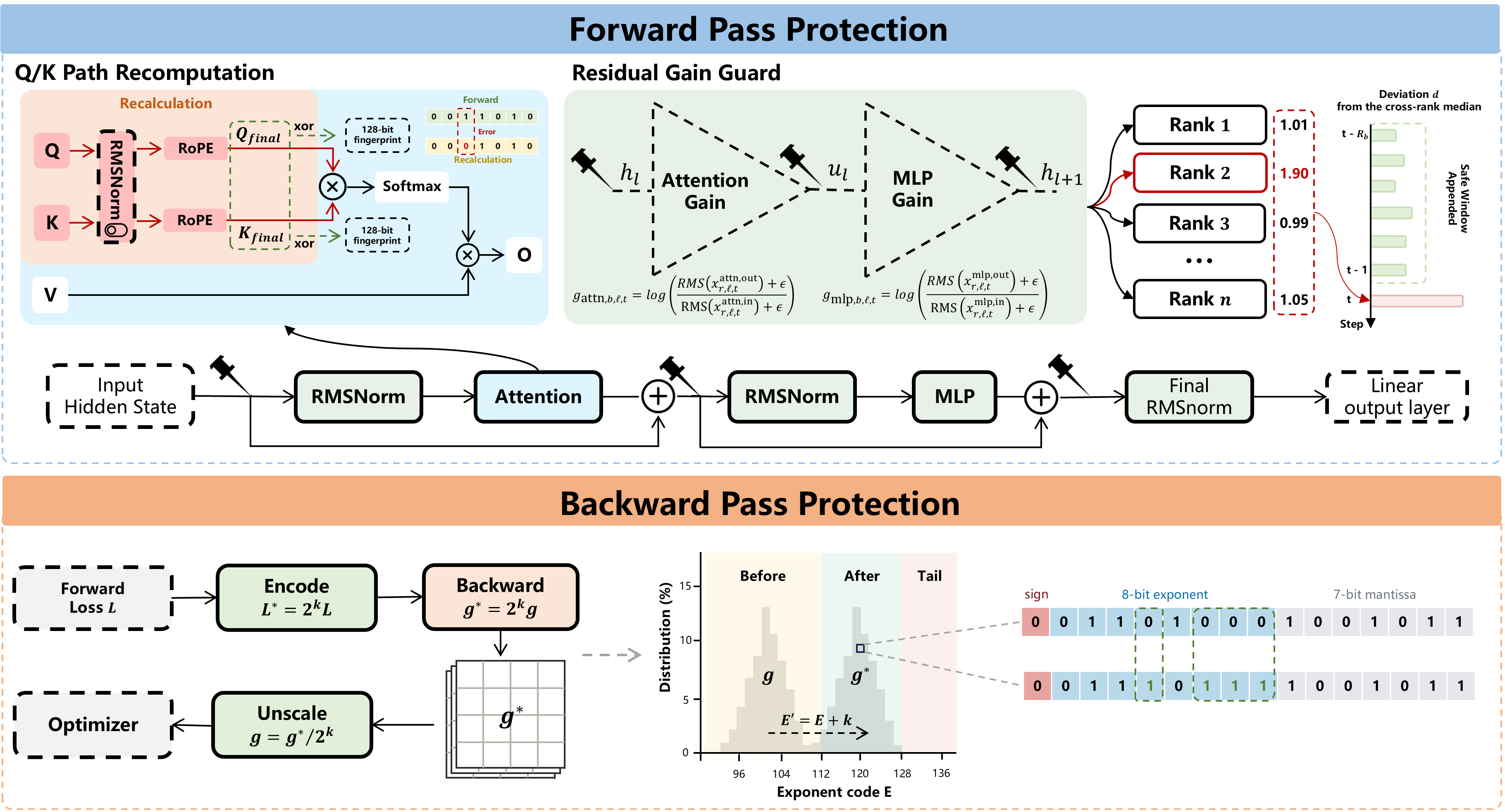}
  \caption{Overview of the protection design. Q/K path recomputation verifies the
Q/K tensors before attention, while the residual gain guard monitors the
residual writebacks after attention and the multilayer perceptron. Exponent
scaling protects gradient computation in the backward pass.}
  \label{fig:method-overview}
\end{figure*}

Figure~\ref{fig:method-overview} illustrates the overall architecture of TrainSDC.
TrainSDC follows the characterization in Section~3 and adopts heterogeneous protection for different stages of Transformer training.
\textbf{For the forward pass}, protection is determined by fault propagation behavior. Computations whose faults become difficult to observe after propagation are protected through direct verification, whereas computations whose faults remain observable are protected through lightweight runtime monitoring.
\textbf{For the backward pass}, vulnerability exhibits much weaker dependence on computation location. TrainSDC therefore employs a unified numerical protection mechanism based on exponent-aware gradient scaling.


\subsection{Forward Pass Protection}
\label{sec:path-aware-forward}



The characterization identifies two categories of forward computations with distinct protection requirements.
Faults in the Q/K path become weakly observable after propagation while causing the most persistent training deviation. These computations therefore require direct verification.
In contrast, faults in the remaining forward computations preserve detectable propagation signatures, allowing them to be protected through lightweight runtime monitoring instead of redundant execution.


\paragraph{Q/K Path Recomputation.}
The preceding observations show that a transient fault at any stage before the
Q/K tensors enter attention can affect training. Protecting only the Q/K
projections is therefore insufficient because faults can also arise during Q/K
normalization or rotary position encoding. We consequently recompute the
complete Q/K path, covering the projections, normalization, and rotary position
encoding. At the same time, under the transient fault model, a
fault is not expected to affect both executions identically, so a mismatch
between them indicates an incorrect Q/K result.

The complete Q/K path is executed twice with the same input and network
parameters. At worker rank $r$, layer $\ell$, and training step $t$, denote the
original tensors entering attention by $Q_{r,\ell,t}$ and $K_{r,\ell,t}$. The
second execution produces $\widehat{Q}_{r,\ell,t}$ and
$\widehat{K}_{r,\ell,t}$ as their recomputed counterparts.

Compact fingerprints avoid retaining and directly comparing the full Q/K
tensors. Each Q or K tensor is reduced to two 64-bit XOR fingerprints, producing
a 128-bit representation. For a flattened tensor
$Z=(z_0,\ldots,z_{n-1})$, let $b_i=\operatorname{bits}(z_i)$ denote the raw
binary encoding of $z_i$, let $w$ denote its bit width, and let $L$ denote the
number of elements packed into one fingerprint word. The fingerprints are
\begin{align}
\phi_{\mathrm{raw}}(Z)
  &= \bigoplus_{i=0}^{n-1}
     \left(b_i \ll w(i\bmod L)\right), \\
\phi_{\mathrm{indexed}}(Z)
  &= \bigoplus_{i=0}^{n-1}
     \operatorname{mix}(i,b_i), \\
\Phi(Z)
  &= \left[
     \phi_{\mathrm{raw}}(Z),
     \phi_{\mathrm{indexed}}(Z)
     \right]^{\top}.
\end{align}
Here, $\oplus$ denotes bitwise XOR, $\ll$ denotes a bit shift, and
$\operatorname{mix}(i,b_i)$ combines each element encoding with its flattened
index. The indexed fingerprint reduces cancellation between changes at
different tensor positions.

A Q/K mismatch is reported when either fingerprint pair differs between the
two executions:
\begin{equation}
a^{QK}_{r,\ell,t}
=
\begin{cases}
1, &
\text{if }\Phi(Q_{r,\ell,t})
\ne\Phi(\widehat{Q}_{r,\ell,t}), \\
1, &
\text{if }\Phi(K_{r,\ell,t})
\ne\Phi(\widehat{K}_{r,\ell,t}), \\
0, & \text{otherwise}.
\end{cases}
\end{equation}

\paragraph{Residual Gain Guard.}

The residual gain guard covers transient faults outside the Q/K path. These faults reach the residual writebacks without
passing through any intermediate normalization, and therefore
arrive as unattenuated amplitude errors. The residual output is then passed
directly to subsequent computation, so the amplitude error propagates through
the network and can be detected from the input-to-output residual gain.

The first step computes the amplitude change across each residual writeback.
For worker rank $r$, layer $\ell$, training step $t$, and
$p\in\{\mathrm{attention},\mathrm{multilayer\ perceptron}\}$, let
$x^{p,\mathrm{in}}_{r,\ell,t}$ and $x^{p,\mathrm{out}}_{r,\ell,t}$ denote the
residual input and output. The root mean square and residual gain are
\begin{align}
R(x)
  &= \sqrt{\frac{1}{|x|}\sum_{i=1}^{|x|}x_i^2}, \\
g^{p}_{r,\ell,t}
  &= \log\!\left(R(x^{p,\mathrm{out}}_{r,\ell,t})+\epsilon\right)
   - \log\!\left(R(x^{p,\mathrm{in}}_{r,\ell,t})+\epsilon\right).
\end{align}
Here, $|x|$ is the number of elements, $R(x)$ represents the endpoint
amplitude, and $g^{p}_{r,\ell,t}$ represents its relative change across the
residual writeback. The logarithm places amplification and attenuation around
zero. The residual gain guard reuses the token-wise mean square computed by
root mean square normalization instead of scanning the residual tensor again.

The second step compares each worker rank with the remaining worker ranks. In
distributed data-parallel training, all worker ranks use the same network
parameters at the same training step, so the remaining worker ranks provide a
current reference. The reference
and rank-specific deviation are
\begin{align}
c^{p}_{-r,\ell,t}
  &= \operatorname*{median}_{j\ne r}g^{p}_{j,\ell,t}, \\
e^{p}_{r,\ell,t}
  &= g^{p}_{r,\ell,t}-c^{p}_{-r,\ell,t}.
\end{align}
Here, $j$ indexes the remaining worker ranks,
$c^{p}_{-r,\ell,t}$ is their median residual gain, and
$e^{p}_{r,\ell,t}$ measures how far worker rank $r$ deviates from that
reference. Excluding worker rank $r$ prevents it from changing its own
reference.

The detection thresholds are calibrated separately for the monitored layers and
residual writebacks. Because normal cross-rank deviations can change with input
data and training progress, a fixed reference may become inaccurate. A sliding
window $\mathcal{H}^{p}_{\ell,t}$ therefore stores recent finite deviations,
with the largest absolute deviation excluded before each update. The normalized
score is

\begin{align}
\mu^{p}_{\ell,t}
  &= \operatorname{median}(\mathcal{H}^{p}_{\ell,t}), \\
s^{p}_{\ell,t}
  &= \max\!\left(
     1.4826
     \operatorname*{median}_{h\in\mathcal{H}^{p}_{\ell,t}}
     |h-\mu^{p}_{\ell,t}|,
     s_{\min}
     \right), \\
z^{p}_{r,\ell,t}
  &= \frac{|e^{p}_{r,\ell,t}-\mu^{p}_{\ell,t}|}
           {s^{p}_{\ell,t}}.
\end{align}

Here, $\mu^{p}_{\ell,t}$ is the typical historical deviation,
$s^{p}_{\ell,t}$ is its median-absolute-deviation scale, and
$z^{p}_{r,\ell,t}$ measures the current deviation relative to normal historical
variation.

After computing $z^{p}_{r,\ell,t}$ for all worker ranks, the residual gain guard
selects the largest score $z_{(1)}$. 
A transient fault is detected when $z_{(1)}$ exceeds the effective threshold. 
The worker rank producing $z_{(1)}$ is identified as the faulty candidate.

\subsection{Backward Pass Protection}
\label{sec:exponent-aware-backward}

Unlike the forward pass, backward vulnerability exhibits little dependence on Transformer modules. This suggests that protecting individual computations separately would incur additional overhead without addressing the primary source of vulnerability.

To identify this common source, we analyze the numerical behavior of gradient representations under bit flips. 
We find that the numerical magnitude of a gradient tensor directly determines its
vulnerability to bit flips. For a bfloat16 value $g$, let $e_j$
denote exponent-field bit $j$. When the original and flipped values are both
finite, flipping $e_j$ changes the magnitude by~\citep{kalamkar2019study}
\begin{equation}
\frac{|\operatorname{flip}_{e_j}(g)|}{|g|}=
\begin{cases}
2^{2^j}, & e_j=0,\\
2^{-2^j}, & e_j=1.
\end{cases}
\label{eq:exponent-flip-gain}
\end{equation}
The effect of a flip therefore depends on the current bit value. Small
gradients occupy low exponent codes, where the selected exponent bits are zero
more often, so a flip amplifies the value. Increasing the magnitude shifts the
exponent-code distribution upward; once the selected bits become one, the same
flip turns one into zero and attenuates the value instead, as illustrated in
Figure~\ref{fig:method-overview}. The two directions cause very different
damage. An amplified corrupted gradient propagates into the global parameter
update, whereas an attenuated corrupted gradient suppresses its own
contribution and causes little deviation. Thus, exponent scaling protects the
backward pass by moving gradients from the amplifying region into the
attenuating region.

Exponent scaling changes exponent codes while preserving the parameter update.
We use $k$ to specify the scaling factor $2^k$. Before the backward pass, the
loss is multiplied by $2^k$, which multiplies every generated gradient by the
same factor and adds $k$ to its exponent code. Immediately before gradient
clipping and the parameter update, the gradient is divided by $2^k$, so the
update is unchanged:
\begin{equation}
g_s=\frac{\partial(2^kL)}{\partial\theta}=2^k g,
\qquad
2^{-k}g_s=g.
\end{equation}

\subsection{Selecting the Scaling Exponent}
\label{sec:scale-selection}
The scaling exponent $k$ is selected by minimizing the predicted vulnerability
within the numerical range of the backward pass. The choice is a trade-off. A
small $k$ leaves many gradients in exponent codes for which a bit flip
amplifies the value, whereas a large $k$ pushes the upper tail of the
distribution toward non-finite values. We therefore collect a histogram $H$ of
gradient exponent codes from clean training steps and evaluate each candidate
by shifting the codes by $k$:
\begin{align}
R_{\mathrm{amp}}(k;H)
  &= \sum_{j\in\mathcal{J}} P_{H,k}(e_j=0)
     \left(2^{2^j}-1\right)^2, \\
k^* &= \arg\min_{k\in\mathcal{K}_{\mathrm{safe}}}
       R_{\mathrm{amp}}(k;H).
\label{eq:nonlinear-exponent-risk}
\end{align}
Here, $\mathcal{J}$ contains the exponent positions included by the fault
model, $P_{H,k}(e_j=0)$ is the fraction of shifted codes for which bit $e_j$
is zero, and $\mathcal{K}_{\mathrm{safe}}$ contains the candidates that keep
the shifted codes within the finite range. Each term weights the probability
of an amplifying flip by its squared relative magnitude change, so
$R_{\mathrm{amp}}$ measures the expected amplification risk of a candidate.
The common bit-selection probability is omitted because it is constant across
the candidates and does not change $k^*$.

\begin{figure}[t]
  \centering
  \includegraphics[width=1.0\columnwidth]{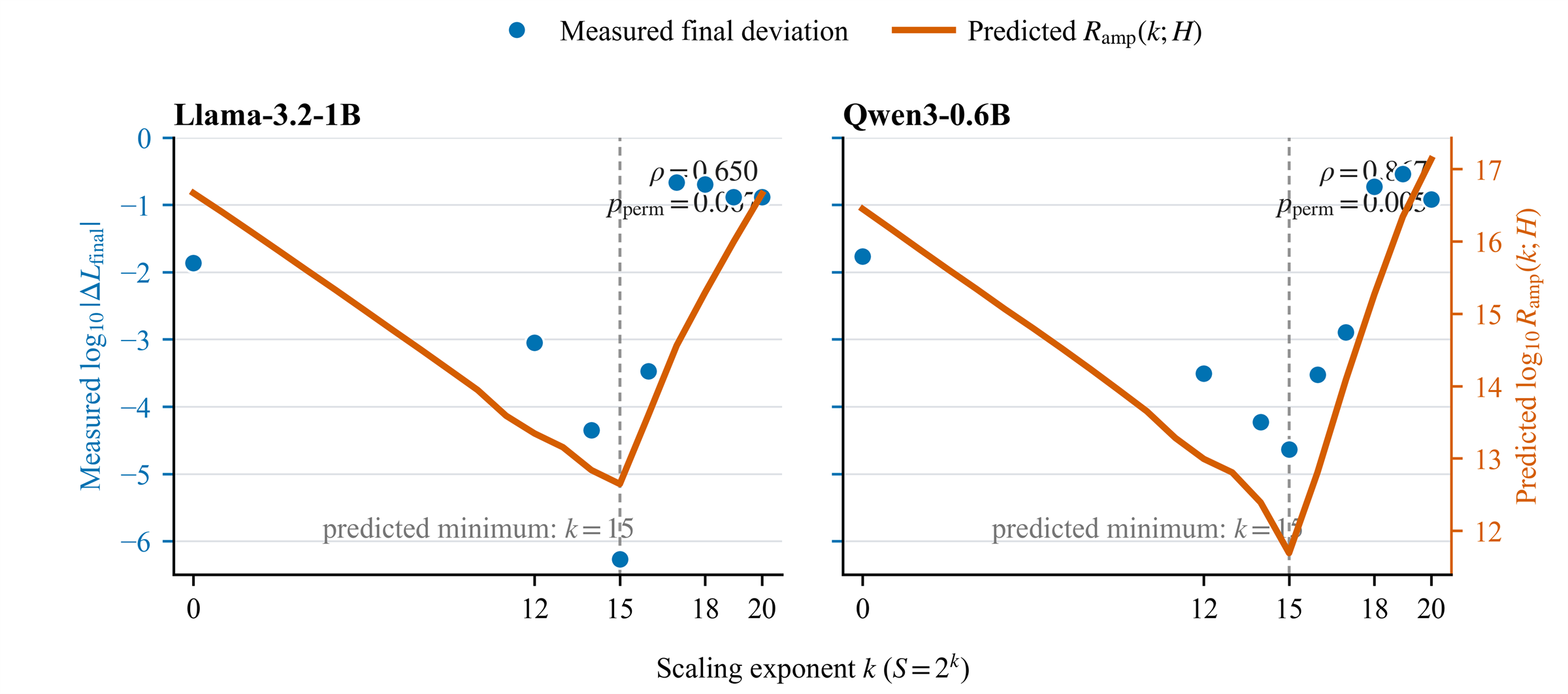}
  \caption{Choosing the scaling factor $k$. The
  measured Final dev. is compared with $R_{\mathrm{amp}}$ across candidate
  values of $k$.}
  \label{fig:backward-scaling}
\end{figure}

Figure~\ref{fig:backward-scaling} compares the measured final deviation with the predicted $R_{\mathrm{amp}}(k;H)$ across the evaluated values of $k$. Spearman's $\rho$ measures whether the two quantities give a consistent ranking of $k$, while $p_{\mathrm{perm}}$ is the exact two-sided permutation probability of observing an association at least this strong under a random ranking. The correlation is $\rho=0.650$ with $p_{\mathrm{perm}}=0.067$ for Llama-3.2-1B and $\rho=0.867$ with $p_{\mathrm{perm}}=0.005$ for Qwen3-0.6B ($n=9$ per network). Based on these results, all subsequent experiments fix the scaling
exponent at the selected optimum $k=15$.

\newcommand{\sdcSci}[2]{\mbox{#1e#2}}

\begin{table*}[t]
  \centering
  {\small
  \setlength{\tabcolsep}{2.5pt}
  \renewcommand{\arraystretch}{1.05}

  \resizebox{\textwidth}{!}{%
  \begin{tabular}{llcccccc}
    \toprule
    \textbf{Network} & \textbf{Method} &
    \makecell{\textbf{Overhead} $\downarrow$\\\textbf{(\%)}} &
    \textbf{$D_{\mathrm{final}}$} $\downarrow$ &
    \textbf{$D_{\mathrm{Maximum}}$} $\downarrow$ &
    \textbf{$\Delta$PPL} $\downarrow$ &
    \makecell{\textbf{Top-1} $\uparrow$\\\textbf{(\%)}} &
    \makecell{\textbf{W L2} $\downarrow$\\\textbf{(\%)}} \\
    \midrule
    \multirow{7}{*}{\makecell[l]{\textbf{Llama 3.2}\\\textbf{1B}}}
    & Fault
    & 0.00 / 0.00
    & \sdcSci{+1.72}{-2} / \textit{nonfinite}
    & \sdcSci{1.23}{-1} / \textit{nonfinite}
    & 4.04 / --
    & 94.20 / --
    & 0.69 / -- \\
    & Loss spike
    & \textbf{$\approx 0$ / $\approx 0$}
    & \sdcSci{+1.72}{-2} / \sdcSci{+1.15}{-1}
    & \sdcSci{1.24}{-1} / \sdcSci{2.84}{-1}
    & 4.03 / 27.60
    & 94.40 / 82.60
    & 0.692 / 2.10 \\
    & Inf/NaN
    & \textbf{$\approx 0$ / $\approx 0$}
    & \sdcSci{+1.72}{-2} / \sdcSci{+1.18}{-1}
    & \sdcSci{1.23}{-1} / \sdcSci{4.28}{-1}
    & 4.04 / 28.60
    & 94.20 / 82.20
    & 0.69 / 2.15 \\
    & ATTNChecker
    & 7.65 / 7.65
    & \sdcSci{+1.75}{-2} / \sdcSci{+1.07}{-1}
    & \sdcSci{3.73}{-2} / \sdcSci{3.86}{-1}
    & 4.07 / 25.40
    & 94.30 / 83.40
    & 0.692 / 2.00 \\
    & Harmful-update
    & 12.10 / 12.10
    & \sdcSci{+1.08}{-2} / \sdcSci{+5.37}{-3}
    & \sdcSci{1.15}{-1} / \sdcSci{1.57}{-1}
    & 2.49 / 1.25
    & 95.00 / 96.90
    & 0.510 / 0.480 \\
    & LLMFT
    & 13.50 / 13.50
    & \sdcSci{+1.68}{-2} / \sdcSci{+9.43}{-2}
    & \sdcSci{9.37}{-2} / \sdcSci{2.42}{-1}
    & 3.86 / 22.20
    & 94.50 / 84.60
    & 0.679 / 1.84 \\
    \addlinespace[1pt]
    & \textbf{Ours}
    & 1.65 / 1.65
    & \textbf{\sdcSci{-3.86}{-5} / \sdcSci{+6.32}{-4}}
    & \textbf{\sdcSci{2.86}{-4} / \sdcSci{9.14}{-3}}
    & \textbf{\sdcSci{-3.37}{-3} / 0.173}
    & \textbf{99.10 / 98.60}
    & \textbf{\sdcSci{2.83}{-2} / 0.149} \\
    \midrule
    \multirow{7}{*}{\makecell[l]{\textbf{Qwen3}\\\textbf{0.6B}}}
    & Fault
    & 0.00 / 0.00
    & \sdcSci{+2.30}{-2} / \textit{nonfinite}
    & \sdcSci{9.65}{-2} / \textit{nonfinite}
    & 5.43 / --
    & 93.10 / --
    & 0.69 / -- \\
    & Loss spike
    & \textbf{$\approx 0$ / $\approx 0$}
    & \sdcSci{+2.17}{-2} / \sdcSci{+1.25}{-1}
    & \sdcSci{9.59}{-2} / \sdcSci{2.82}{-1}
    & 5.24 / 36.30
    & 93.60 / 81.80
    & 0.677 / 2.60 \\
    & Inf/NaN
    & \textbf{$\approx 0$ / $\approx 0$}
    & \sdcSci{+2.30}{-2} / \sdcSci{+1.27}{-1}
    & \sdcSci{9.65}{-2} / \sdcSci{3.21}{-1}
    & 5.43 / 36.90
    & 93.10 / 81.70
    & 0.690 / 2.67 \\
    & ATTNChecker
    & 36.70 / 36.70
    & \sdcSci{+2.10}{-2} / \sdcSci{+1.26}{-1}
    & \sdcSci{2.67}{-2} / \sdcSci{2.93}{-1}
    & 5.06 / 37.10
    & 93.70 / 81.70
    & 0.656 / 2.69 \\
    & Harmful-update
    & 11.20 / 11.20
    & \sdcSci{+1.39}{-2} / \sdcSci{+3.82}{-3}
    & \sdcSci{8.85}{-2} / \sdcSci{1.56}{-1}
    & 3.24 / 1.14
    & 94.50 / 97.90
    & 0.471 / 0.361 \\
    & LLMFT
    & 5.62 / 5.62
    & \sdcSci{+2.17}{-2} / \sdcSci{+1.05}{-1}
    & \sdcSci{7.06}{-2} / \sdcSci{2.48}{-1}
    & 5.33 / 29.50
    & 93.60 / 84.20
    & 0.676 / 2.25 \\
    \addlinespace[1pt]
    & \textbf{Ours}
    & 6.76 / 6.76
    & \textbf{\sdcSci{+3.87}{-6} / \sdcSci{+6.15}{-4}}
    & \textbf{\sdcSci{2.22}{-4} / \sdcSci{1.48}{-2}}
    & \textbf{\sdcSci{-1.25}{-3} / 0.175}
    & \textbf{99.10 / 98.80}
    & \textbf{\sdcSci{2.16}{-2} / 0.109} \\
    \bottomrule
  \end{tabular}%
  }
  }

  \caption{Evaluation under the transient-fault injection protocol. Each slash-separated metric reports results for 10 corrupted activation elements (left) and 100{,}000 corrupted activation elements (right) per injection event.}
  \label{tab:method-comparison}
\end{table*}

\begin{table}[t]
  \centering
  \captionsetup{width=\columnwidth,justification=justified,singlelinecheck=false}

  {\small
  \setlength{\tabcolsep}{2.0pt}
  \renewcommand{\arraystretch}{1.05}
  \resizebox{0.9\columnwidth}{!}{%

  \begin{tabular}{@{}llcc@{}}
    \toprule
    \textbf{Network} & \textbf{Variant} &
    \textbf{$D_{\mathrm{final}}$ $\downarrow$} & \textbf{$\Delta$PPL $\downarrow$} \\
    \midrule
    \multirow{4}{*}{Llama}
      & Q/K path Only
      & \sdcSci{+1.69}{-2} / \sdcSci{+9.03}{-2}
      & 4.00 / 21.70 \\
      & \makecell[l]{Residual Only}
      & \sdcSci{+1.44}{-2} / \sdcSci{+9.02}{-2}
      & 3.37 / 21.40 \\
      & \makecell[l]{Exponent scaling Only}
      & \sdcSci{+3.03}{-4} / \textit{Inf}
      & \sdcSci{7.28}{-2} / -- \\
      & \textbf{Complete}
      & \textbf{\sdcSci{-3.86}{-5} / \sdcSci{+6.32}{-4}}
      & \textbf{\sdcSci{-3.37}{-3} / 0.173} \\
    \midrule
    \multirow{4}{*}{Qwen}
      & Q/K path Only
      & \sdcSci{+2.15}{-2} / \sdcSci{+1.02}{-1}
      & 5.22 / 28.50 \\
      & \makecell[l]{Residual Only}
      & \sdcSci{+1.98}{-2} / \sdcSci{+1.04}{-1}
      & 4.59 / 29.10 \\
      & \makecell[l]{Exponent scaling Only}
      & \sdcSci{+1.12}{-3} / \textit{Inf}
      & 0.250 / -- \\
      & \textbf{Complete}
      & \textbf{\sdcSci{+3.87}{-6} / \sdcSci{+6.15}{-4}}
      & \textbf{\sdcSci{-1.25}{-3} / 0.175} \\
    \bottomrule
  \end{tabular}
    }
  }
  \caption{Component ablation under the unified injection protocol. Each slash-separated metric reports results for 10 corrupted activation elements (left) and
100{,}000 corrupted activation elements (right) per injection event.}
  \label{tab:ablation-study}
  \vspace{2pt}

  \parbox{\columnwidth}{\footnotesize

  }
\end{table}

\section{Evaluation}
\label{sec:experiments}

\providecommand{\todo}[1]{TBD}

\subsection{Experimental Setup}

\paragraph{Models and training.}
The evaluation uses Llama~3.2-1B (16 layers) \citep{grattafiori2024llama} and
Qwen3-0.6B (28 layers) \citep{yang2025qwen3}, instantiated from their
configurations with random weights (seed 8) and without pretrained weights.
Every fault-free, faulty, protected, and ablation run resumes from a common
step-1221 checkpoint and ends at step 2035. Training uses data parallelism on six
NVIDIA A100 80-GB graphics processing units, one 1024-token sequence per rank,
and eight accumulated microbatches, for 49,152 tokens per network update.
Computation uses bfloat16 automatic mixed precision and float32 parameters.
AdamW uses $(\beta_1,\beta_2)=(0.9,0.95)$, $\epsilon=10^{-8}$, weight decay 0.1,
and global gradient clipping at 1.0. The learning rate warms up for 100 steps to
$5\times10^{-5}$ and cosine-decays to 10\% of that value.

\paragraph{Data.}
The evaluation uses the same document-disjoint SmolLM2-derived split for both
networks \citep{allal2025smollm2}. The split contains 1,193,639 training and
12,056 evaluation documents sampled proportionally from DCLM-Edu, FineWeb-Edu,
Stack-Edu, InfiMM-WebMath, FineMath, and Cosmopedia-v2, then tokenized separately
for each network. The available training and evaluation corpora contain
1.102B/10.959M Llama tokens and 1.126B/11.181M Qwen tokens. Each run consumes
about 100M training tokens, including 40M in the aligned continuation. Final
evaluation covers 614,400 held-out token positions, and no evaluation document
occurs in training.

\paragraph{Fault injection.}
We instantiate the fault model of Section~\ref{sec:fault-model} as follows. To additionally evaluate detection capability under sparse faults, we introduce
a 10-element injection setting. 
For each spatial intensity, the protocol
triggers 813 injection events; all
remaining settings are identical.

\paragraph{Protection configuration.}
The protection configuration applies fused recomputation to the Q/K path,
the residual gain guard to residual writebacks, and exponent scaling to the
backward pass. Fused recomputation covers the Q/K path in every layer, while the
residual gain guard samples layers $\{0,5,10,15\}$ in Llama and
$\{0,9,18,27\}$ in Qwen. An alarm discards the accumulated gradients and replays
all accumulated microbatches without a transient fault before the network update.
Exponent scaling uses $k=15$, and automatic scale adjustment is disabled. 

\paragraph{Metrics.}
The evaluation reports Overhead, $D_{\mathrm{final}}$, $D_{\mathrm{Maximum}}$, $\Delta$PPL, Top-1 and W L2. $D_{\mathrm{final}}$ and $D_{\mathrm{Maximum}}$ follow
Section~\ref{sec:fault-characterization}, and $\Delta$PPL is the held-out
perplexity difference from the aligned fault-free run. Top-1 measures token agreement, W L2 measures
relative parameter drift. Overhead is the change in mean fault-free step time
over 100 measured optimizer steps. 

\subsection{Main Results}

All methods use the same checkpoints, data order, and injection schedules.
Harmful-update follow prior training studies
\citep{altenbernd2026exploring}, and ATTNChecker follows its
attention-protection setting \citep{liang2025attnchecker}.
LLMFT uses its learned detector through our zero-shot oracle-route adapter
\citep{yu2025exploring}. Slash-separated values denote 10 / 100,000 corrupted
elements.

At 10 elements, the complete method achieves near-zero $D_{\mathrm{final}}$, $D_{\mathrm{Maximum}}$
below $2.9\times10^{-4}$, and negligible $\Delta$PPL on both models, with
1.65\% overhead on Llama and 6.76\% on Qwen. At 100,000 elements, unprotected
training diverges, whereas the complete method preserves convergence consistent
with fault-free training. Its $D_{\mathrm{final}}$ is $6.32\times10^{-4}$ /
$6.15\times10^{-4}$, $D_{\mathrm{Maximum}}$ is $9.14\times10^{-3}$ /
$1.48\times10^{-2}$, and $\Delta$PPL is 0.173 / 0.175. It achieves the best
reported outcome metrics on both models and at both fault intensities.

\subsection{Ablation Study}

The ablation evaluates the Q/K-path check, residual gain guard, and exponent
scaling individually, with the complete method included as a reference. All
variants use the same injection protocol as Table~\ref{tab:method-comparison}.

At 10 elements, exponent scaling is the strongest individual component,
reducing $D_{\mathrm{final}}$ to $3.03\times10^{-4}$ / $1.12\times10^{-3}$ and
$\Delta$PPL to 0.0728 / 0.250 for Llama/Qwen. At 100,000 elements, K15-only
encounters \texttt{found\_inf}, so its $D_{\mathrm{final}}$ and PPL are not reported.
Q/K-only and residual-only produce $D_{\mathrm{final}}$ near $10^{-1}$ and $\Delta$PPL
between 21.4 and 29.1, whereas the complete method reduces $D_{\mathrm{final}}$ to
$6.32\times10^{-4}$ / $6.15\times10^{-4}$ and $\Delta$PPL to 0.173 / 0.175.
The three components therefore provide complementary protection under strong
faults.

\section{Conclusion}

This work shows that query/key path faults cause more persistent damage, whereas
value, attention-output, and multilayer-perceptron faults cause larger but
shorter-lived loss changes. Backward vulnerability varies less across
components and depends on the gradient exponent distribution. Based on these
findings, TrainSDC combines query/key path recomputation, the Residual Gain
Guard, and exponent scaling. On Llama~3.2-1B and Qwen3-0.6B, TrainSDC mitigates
both sparse and dense faults with limited overhead and restores training
outcomes to a level close to fault-free training.

\bibliographystyle{plainnat}
\bibliography{main}

\clearpage
\onecolumn
\appendix

\providecommand{\appfill}[1]{\textbf{[FILL: #1]}}
\providecommand{\appCleanSteps}{200}
\providecommand{\apptablestyle}{%
  \small
  \setlength{\tabcolsep}{5pt}%
  \renewcommand{\arraystretch}{1.12}%
}

\section{Appendix}
\label{app:additional-results}

\subsection{Robustness across Fault Rate, Layer, and Injection Budget}
\label{app:layer-budget-robustness}

\paragraph{Motivation.}
The main characterization of SmolLM2 uses Layer 20, injects faults in 1\% of
training steps, and corrupts 100,000 elements in each selected tensor.  We test
whether its forward-pass observations remain visible when varying the fault-step
rate, target layer, and injection budget separately.

\begin{figure}[H]
\centering
\includegraphics[width=\linewidth]{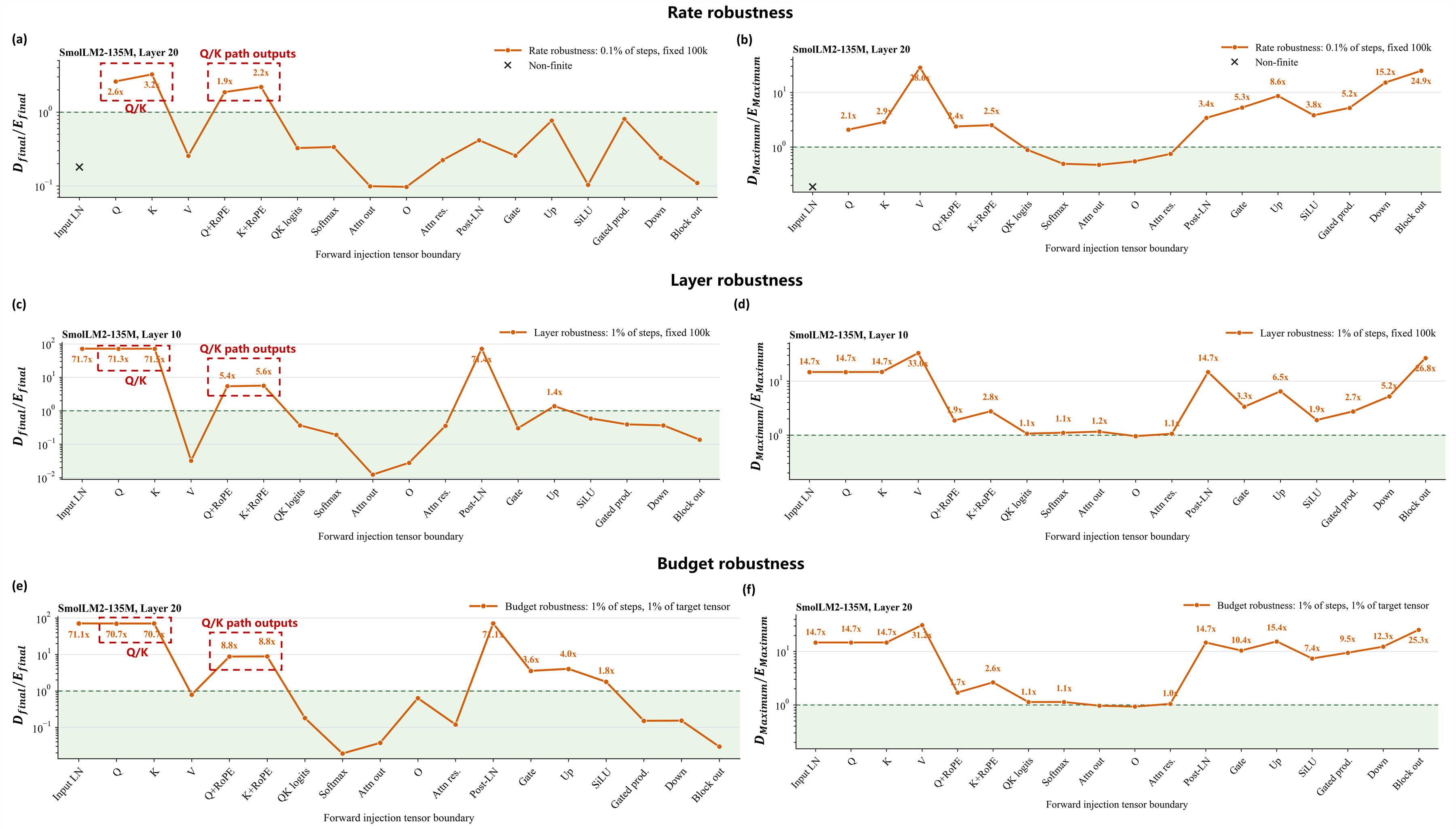}
\caption{Additional forward-pass characterization of SmolLM2-135M.  Panels
(a) and (b) evaluate rate robustness at Layer 20 with a 0.1\% fault-step rate
and 100,000 corrupted elements.  Panels (c) and (d) evaluate layer robustness
at Layer 10 with a 1\% fault-step rate and 100,000 corrupted elements.  Panels
(e) and (f) evaluate budget robustness at Layer 20 with a 1\% fault-step rate
and exactly 1\% of each target tensor corrupted. }
\label{fig:app-layer-budget}
\end{figure}

\paragraph{Experimental variants.}
Figure~\ref{fig:app-layer-budget} reports three controlled variants on
SmolLM2-135M.  Rate robustness uses Layer 20, a 0.1\% fault-step rate, and
100,000 corrupted elements.  Layer robustness uses Layer 10, a 1\% fault-step
rate, and 100,000 corrupted elements.  Budget robustness uses Layer 20, a 1\%
fault-step rate, and corrupts exactly 1\% of each target tensor.  For a target
tensor $x_i$ containing $N_i$ elements, the last variant uses
\begin{equation}
    m_i = \operatorname{round}(0.01 N_i)
\end{equation}
corrupted elements.  All remaining settings, including the
training continuation, temporal injection schedule, random seeds, bit-flip
policy, optimizer, and fault-free reference trajectory, follow the main
characterization.

\paragraph{Results.}
Across all three variants, forward sensitivity remains path-dependent.
Q/K-path faults consistently leave final deviations above the fault-free
envelope, while normalization outputs are also vulnerable.  In contrast, V,
multilayer perceptron, and block-output faults mainly produce large peak
deviations that are attenuated during subsequent training.  These results agree
with Observation O1 and show that the forward-pass pattern remains under changes
in fault-step rate, target layer, and injection budget.

\FloatBarrier
\subsection{Operating Characteristics of Forward Protection}
\label{app:detector-operating-characteristics}



\paragraph{Threshold calibration and online state.}
Calibration starts from the common step-1221 checkpoint and runs 100
fault-free optimizer steps (steps 1222--1321).  The sliding-window length is
$W=64$ accepted detector observations, normally one per optimizer step.  For
each monitored layer and residual path, one clean observation contributes the
$R-1$ rank deviations with the smallest absolute values; the largest rank
deviation is omitted.  With $R=6$, the stored window
therefore contains $64\times5=320$ scalar deviations per layer and path.  The
first 64 calibration steps seed this window, while all 100 calibration steps
are used to determine the calibrated threshold floors.  Consequently, the
evaluation run begins with a full window and has no additional deployment
warm-up. 

The robust scale follows the definition in the main text.  The detector uses
threshold coefficient $\alpha$.  Let $c^{p}_{\mathrm{rel},\ell,\mathrm{cal}}$ and
$c^{p}_{z,\ell,\mathrm{cal}}$ denote the largest eligible relative-deviation
and robust-$z$ scores observed in the 100 clean calibration steps before any
safety scaling.  With $s_{\min}=10^{-6}$ and adaptive quantile $q=1$, the
effective thresholds at step $t$ are
\begin{align}
\tau^{p}_{\mathrm{rel},\ell,t}
  &= \alpha\max\!\left(
       c^{p}_{\mathrm{rel},\ell,\mathrm{cal}},
       \max_{h\in\mathcal{H}^{p}_{\ell,t}}
       \left(\exp(|h|)-1\right)
     \right), \\
\tau^{p}_{z,\ell,t}
  &= \alpha\max\!\left(
       c^{p}_{z,\ell,\mathrm{cal}},
       \max_{h\in\mathcal{H}^{p}_{\ell,t}}
       \frac{|h-\operatorname{median}(\mathcal{H}^{p}_{\ell,t})|}
       {\max(1.4826\operatorname{MAD}(\mathcal{H}^{p}_{\ell,t}),10^{-6})}
     \right).
\end{align}
Thus, the same coefficient jointly scales the fixed calibration floor and the
adaptive history term.  The floor prevents an unusually quiet recent window
from making the detector over-sensitive, while the history term follows
legitimate training drift.

\paragraph{Selection of $\alpha$.}
We sweep $\alpha\in\{1,1.5,2,2.5,3\}$ using the same four monitored layers,
checkpoint, detector history, fault locations, and random seed.  For each
candidate, clean false positives are measured on 200 held-out fault-free
optimizer steps.  Detection is measured on 96 residual-path snapshot-replay
events per network: 48 events with 10 corrupted elements and 48 events with
100,000 corrupted elements.  Table~\ref{tab:app-rgg-alpha-sweep} reports the
result.

\begin{table}[H]
\centering
\apptablestyle
\begin{tabular}{lrrrr}
\toprule
Network & $\alpha$ & Clean step FPR & 10-element recall & 100,000-element recall \\
\midrule
Llama 3.2-1B & 1.0 & 33/200 (16.5\%) & 48/48 (100.00\%) & 48/48 (100.00\%) \\
              & 1.5 &  3/200 (1.5\%)  & 17/48 (35.42\%)  & 41/48 (85.42\%) \\
              & 2.0 &  1/200 (0.5\%)  & 17/48 (35.42\%)  & 41/48 (85.42\%) \\
              & \textbf{2.5} & \textbf{0/200 (0\%)} & \textbf{17/48 (35.42\%)} & \textbf{41/48 (85.42\%)} \\
              & 3.0 &  0/200 (0\%)    & 17/48 (35.42\%)  & 41/48 (85.42\%) \\
\midrule
Qwen3-0.6B   & 1.0 & 17/200 (8.5\%)  &  7/48 (14.58\%)  & 32/48 (66.67\%) \\
              & 1.5 &  0/200 (0\%)    &  7/48 (14.58\%)  & 32/48 (66.67\%) \\
              & 2.0 &  0/200 (0\%)    &  7/48 (14.58\%)  & 32/48 (66.67\%) \\
              & \textbf{2.5} & \textbf{0/200 (0\%)} & \textbf{7/48 (14.58\%)} & \textbf{32/48 (66.67\%)} \\
              & 3.0 &  0/200 (0\%)    &  7/48 (14.58\%)  & 32/48 (66.67\%) \\
\bottomrule
\end{tabular}
\caption{Selection of the single effective Residual Gain Guard coefficient
$\alpha$.  Clean FPR is the optimizer-step false-positive rate on 200 held-out
fault-free steps.  The two recall columns report detected events out of 48.}
\label{tab:app-rgg-alpha-sweep}
\end{table}

The setting $\alpha=1$ is over-sensitive on both networks.  From
$\alpha=1.5$ onward, recall is unchanged on this sweep, while the clean FPR
continues to decrease.  We select $\alpha=2.5$ because it is the smallest
candidate with zero held-out clean alarms on both networks and shows no recall
reduction relative to $\alpha=1.5$ or $2$.  The resulting per-layer,
per-path calibration floors are listed in
Table~\ref{tab:app-sparse-rgg-thresholds}.

\begin{table}[H]
\centering
\apptablestyle
\setlength{\tabcolsep}{4pt}
\begin{tabular}{llrrrr}
\toprule
Network & Layer
& $\tau_{\mathrm{rel}}^{\mathrm{Attn.}}$
& $\tau_z^{\mathrm{Attn.}}$
& $\tau_{\mathrm{rel}}^{\mathrm{MLP}}$
& $\tau_z^{\mathrm{MLP}}$ \\
\midrule
Llama 3.2-1B & 0  & 0.340077 & 10.926 & 0.322114 & 11.736 \\
              & 5  & 0.008262 & 11.462 & 0.017856 & 25.025 \\
              & 10 & 0.008631 & 16.026 & 0.019817 & 13.107 \\
              & 15 & 0.007139 & 14.899 & 0.033000 &  9.942 \\
\midrule
Qwen3-0.6B   & 0  & 0.282147 & 11.798 & 0.206769 & 13.587 \\
              & 9  & 0.008335 & 15.364 & 0.011517 & 19.679 \\
              & 18 & 0.006522 & 15.650 & 0.009818 & 16.609 \\
              & 27 & 0.010103 & 13.767 & 0.018967 & 12.445 \\
\bottomrule
\end{tabular}
\caption{Per-layer, per-path calibrated threshold floors after applying the
selected $\alpha=2.5$.  ``Attn.'' and ``MLP'' denote the attention and
multilayer-perceptron residual writebacks.}
\label{tab:app-sparse-rgg-thresholds}
\end{table}

\paragraph{Alarm aggregation and history admission.}
For each layer and path, the implementation considers the largest and
second-largest rank scores.  The robust-$z$ branch alarms when the largest
score reaches $\tau_z$ and is at least $1.5$ times the second-largest score; a
companion relative-deviation branch applies the same dominance rule using
$\tau_{\mathrm{rel}}$.  A nonfinite or negative reused mean-square value bypasses
both thresholds and causes an immediate alarm.  Nonfinite values are never
written to the history.

All rank, path, and monitored-layer decisions are OR-reduced into one
optimizer-step alarm.  Multiple simultaneous alarms retain their individual
diagnostics but trigger only one replay of the accumulated microbatches, with
fault injection disabled, before a single optimizer update.  An attempt rejected
by the Residual Gain Guard is not written to its window.

The online window persists within an uninterrupted training process but is not
stored in the ordinary model/optimizer checkpoint.  On restart, the reported
implementation reloads the same 64-step seed from the separate calibration
file rather than carrying recent deployment history across the checkpoint.
All reported main runs were uninterrupted, so this restart behavior does not
alter their measurements.

\paragraph{Results.}
Table~\ref{tab:app-rgg-alpha-sweep} reports the held-out clean behavior used to
select $\alpha$.  At the selected $\alpha=2.5$, neither network raises an alarm
in the 200-step clean holdout.
We then run an independent 814-step fault-free deployment trace and
three complete 813-event fault traces per network.  The complete clean traces
produce zero alarms for Qwen and one alarm for Llama, corresponding to a
combined step-level FPR of $1/1628=0.061\%$.  This longer measurement is the
clean FPR reported below; it does not change the held-out sweep used to select
$\alpha$.

\begin{table}[H]
\centering
\apptablestyle
\setlength{\tabcolsep}{5pt}
\begin{tabular}{lccccc}
\toprule
Network & Clean alarms & 10 elements & 1,000 elements & 100,000 elements
& All fault runs \\
\midrule
Llama 3.2-1B
& 1/814 (0.12\%)
& 175/403 (43.42\%)
& 338/403 (83.87\%)
& 348/403 (86.35\%)
& 861/1209 (71.22\%) \\
Qwen3-0.6B
& 0/814 (0\%)
& 165/403 (40.94\%)
& 279/403 (69.23\%)
& 316/403 (78.41\%)
& 760/1209 (62.86\%) \\
\bottomrule
\end{tabular}
\caption{Complete-run clean FPR and forward-alarm recall with $\alpha=2.5$.
Each fault run contains 403 valid forward injection events.  An event is
detected when Q/K Path Recomputation or the Residual Gain Guard requests a
pre-update replay.}
\label{tab:app-detector-operating-point}
\end{table}

\begin{table}[H]
\centering
\apptablestyle
\begin{tabular}{lrrrrr}
\toprule
Injection module or node & 10 elements & 1,000 elements & 100,000 elements
& Total & Share (\%) \\
\midrule
Attention weights             & 45 & 45 & 45 & 135 & 16.94 \\
Attention logits              & 32 & 32 & 32 &  96 & 12.05 \\
Down projection               & 39 & 28 & 20 &  87 & 10.92 \\
Attention residual            & 28 & 28 & 27 &  83 & 10.41 \\
Pre-$o$ attention output      & 42 & 17 &  8 &  67 &  8.41 \\
Output projection             & 25 & 21 & 10 &  56 &  7.03 \\
Input RMSNorm                 & 48 &  0 &  0 &  48 &  6.02 \\
SiLU output                   & 37 &  7 &  0 &  44 &  5.52 \\
Up projection                 & 35 &  1 &  0 &  36 &  4.52 \\
Gate projection               & 32 &  3 &  0 &  35 &  4.39 \\
Post-attention RMSNorm        & 33 &  0 &  0 &  33 &  4.14 \\
Gated MLP product             & 27 &  3 &  0 &  30 &  3.76 \\
Value projection              & 25 &  3 &  0 &  28 &  3.51 \\
Block output                  & 18 &  1 &  0 &  19 &  2.38 \\
\midrule
All protected Q/K-path targets & 0 & 0 & 0 & 0 & 0.00 \\
\bottomrule
\end{tabular}
\caption{Post-hoc composition of missed forward alarms with $\alpha=2.5$,
aggregated over both networks.  Share is relative to all 797 missed forward
events across the three spatial intensities.}
\label{tab:app-forward-miss-composition}
\end{table}

Across all three intensities, Q/K Path Recomputation detects every direct
corruption of the protected Q/K-path targets.  The remaining misses are
concentrated in attention weights, attention logits, and residual or projection
outputs outside that exact-recomputation boundary.  Increasing the number of
corrupted elements primarily improves Residual Gain Guard recall: the combined
recall rises from 43.42\% to 86.35\% for Llama and from 40.94\% to 78.41\% for
Qwen.



\FloatBarrier
\subsection{Temporal Stability of the Exponent-Scaling Parameter}
\label{app:k-stability}

\paragraph{Motivation.}
Exponent scaling selects $k$ from the exponent-code histogram of clean backward
gradients.  This selection is practical only if the same value can be held for
a substantial training interval.  We therefore measure how the selected value
evolves over the complete public checkpoint trajectory of Pythia-1B.

\paragraph{Checkpoint sweep.}
We evaluate all 154 released checkpoints from step 0 through step 143,000.  At
each checkpoint, we run 12 fixed WikiText-103 validation sequences with batch
size 1 and sequence length 1,024.  We collect the BF16 exponent histogram $H_t$
of $\mathrm{d}L/\mathrm{d}Y$ at all 64 Transformer linear projections, using
8,192 deterministic samples per projection and sequence, or 6,291,456 samples
per checkpoint.  No fault is injected during profiling.  We evaluate the same
candidate set as in the main method and compute the pointwise selection
\begin{equation}
    k^{*}(t)
    = \operatorname*{arg\,min}_{k\in\mathcal{K}_{\mathrm{safe}}}
      R_{\mathrm{amp}}(k;H_t).
\end{equation}
For phase-wise deployment, a held value $k_{\mathrm{held}}$ remains admissible
while its predicted risk is at most 25\% above the pointwise minimum.  A new
profile is required only when this condition no longer holds.  The reported
values of $k$ use the same training-equivalent convention as the main
experiments.

\begin{figure}[H]
\centering
\includegraphics[width=0.90\linewidth]{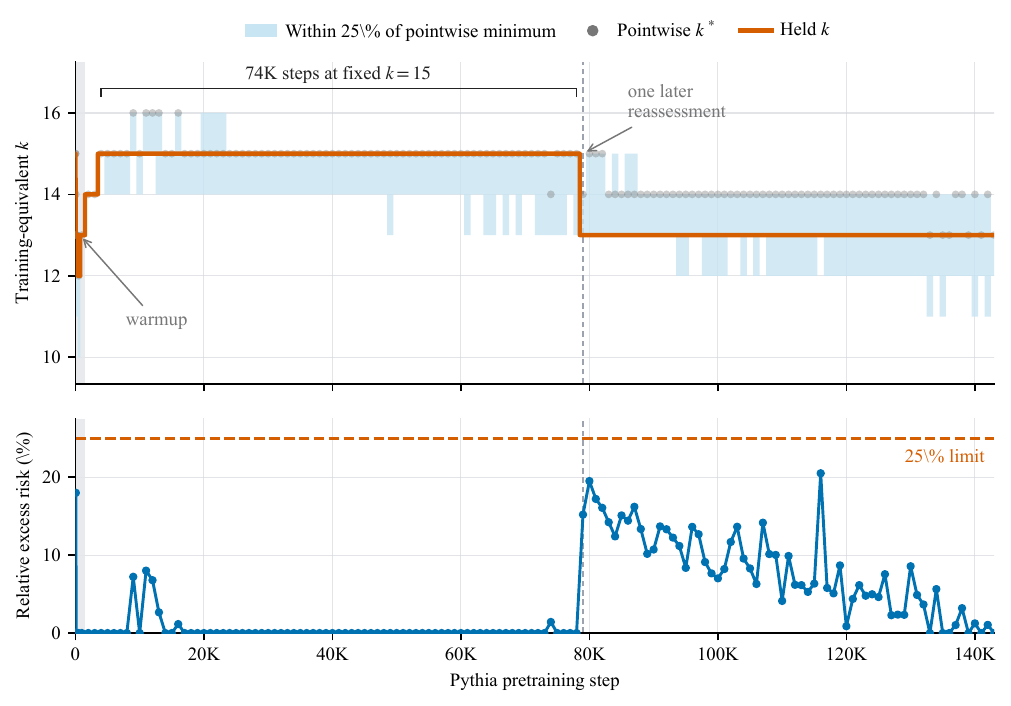}
\caption{Temporal stability of exponent scaling across all 154 public
Pythia-1B checkpoints.  The upper panel shows the pointwise optimum $k^{*}$,
the values within 25\% of its predicted risk, and the value held for each
phase.  The lower panel shows the relative excess risk of the held value.  The
shaded region marks warm-up, and the vertical dashed line marks the only later
reassessment after the initial post-warm-up value is established at step
4,000.}
\label{fig:app-pythia-k}
\end{figure}

\paragraph{Results.}
Figure~\ref{fig:app-pythia-k} shows frequent movement during warm-up and the
immediate post-warm-up transition.  Once $k=15$ is selected at step 4,000, it
remains admissible through step 78,000, a span of 74,000 training steps.  One
later reassessment at step 79,000 selects $k=13$, which remains admissible
through the final checkpoint at step 143,000.  The relative excess risk remains
below the 25\% limit throughout both intervals.  Thus, after the initial
post-warm-up operating point is established, the remaining trajectory of more
than 100,000 steps requires only one additional reassessment.

\paragraph{Practical implication.}
The scaling exponent $k$ is evaluated more frequently
during warm-up, when the histogram changes rapidly, and is then held fixed over
long training intervals.  For this trajectory, the initial post-warm-up value
requires only one reassessment in the latter half of training. 

\FloatBarrier

\end{document}